\pdfoutput=1
\let\accentvec\vec
\documentclass[runningheads]{llncs}

\let\vec\accentvec
\usepackage{graphicx}
\usepackage{amsmath,amssymb} % define this before the line numbering.
\usepackage{color}
\usepackage[width=122mm,left=12mm,paperwidth=146mm,height=193mm,top=12mm,paperheight=217mm]{geometry}

\usepackage{float}
\usepackage[linesnumbered, boxed, ruled]{algorithm2e}
\usepackage{multirow}
\usepackage{array}
\newcolumntype{C}[1]{>{\centering\let\newline\\\arraybackslash\hspace{0pt}}m{#1}}
\usepackage{booktabs}
\usepackage{longtable}
\newcommand{\repeatthanks}{\textsuperscript{\thefootnote}}

\begin{document}
\pagestyle{headings}
\mainmatter

\title{SRDA: Generating Instance Segmentation Annotation Via Scanning, Reasoning And Domain Adaptation}
% Replace with your title

\titlerunning{SRDA}
% Replace with a meaningful short version of your title

\authorrunning{Wenqiang Xu, Yonglu Li and Cewu Lu}
% Replace with shorter version of the author list. If there are more authors than fits a line, please use A. Author et al.

\author{Wenqiang Xu\thanks{these two authors contributed equally}, Yonglu Li\repeatthanks, Cewu Lu}

\institute{Department of Computer Science and Engineering,\\
	Shanghai Jiaotong University\\
	\email{ \{vinjohn,yonglu\_li,lucewu\}@sjtu.edu.cn}
}

\maketitle

\begin{abstract}
Instance segmentation is a problem of significance in computer vision. However, preparing annotated data for this task is extremely time-consuming and costly. By combining the advantages of 3D scanning, reasoning, and GAN-based domain adaptation techniques, we introduce a novel pipeline named SRDA to obtain large quantities of training samples with very minor effort. Our pipeline is well-suited to scenes that can be scanned, i.e. most indoor and some outdoor scenarios. To evaluate our performance, we build three representative scenes and a new dataset, with 3D models of various common objects categories and annotated real-world scene images. Extensive experiments show that our pipeline can achieve decent instance segmentation performance given very low human labor cost.

\end{abstract}

\section{Introduction}
Instance segmentation \protect\cite{MdNC,YiLi2017} is one of the fundamental problems in computer vision, which provides many more details in comparison to object detection \protect\cite{renNIPS15fasterrcnn}, or semantic segmentation \protect\cite{Long2015Fully}. With the development of deep learning, significant progress has been made in instance segmentation. Many annotated datasets of large quantity were proposed \protect\cite{Cordts2016Cityscapes,Lin2014Microsoft}. However, in practice, when meeting a new environment with many new objects, large-scale training data collection and annotation is inevitable, which is cost-prohibitive and time-consuming.

Researchers have longed for a means of generating numerous training samples with minor effort.
Computer graphics simulation is a promising way, since a 3D scene can be a source of unlimited photorealistic images paired with ground truths. Besides, modern simulation techniques are capable of synthesizing most indoor and outdoor scenes with perceptual plausibility. Nevertheless, these two advantages are double-edged, rendered images would be painstaking to make the simulated scene visually realistic \protect\cite{Zhu2016Target,Tzeng2015Towards,Rusu2016Sim}. Moreover, for new environment, it is very likely some of the objects in reality are not in the 3D model database. 

\begin{figure}[!ht]
	\begin{center}
		\includegraphics[width=8.3cm,height=3cm]{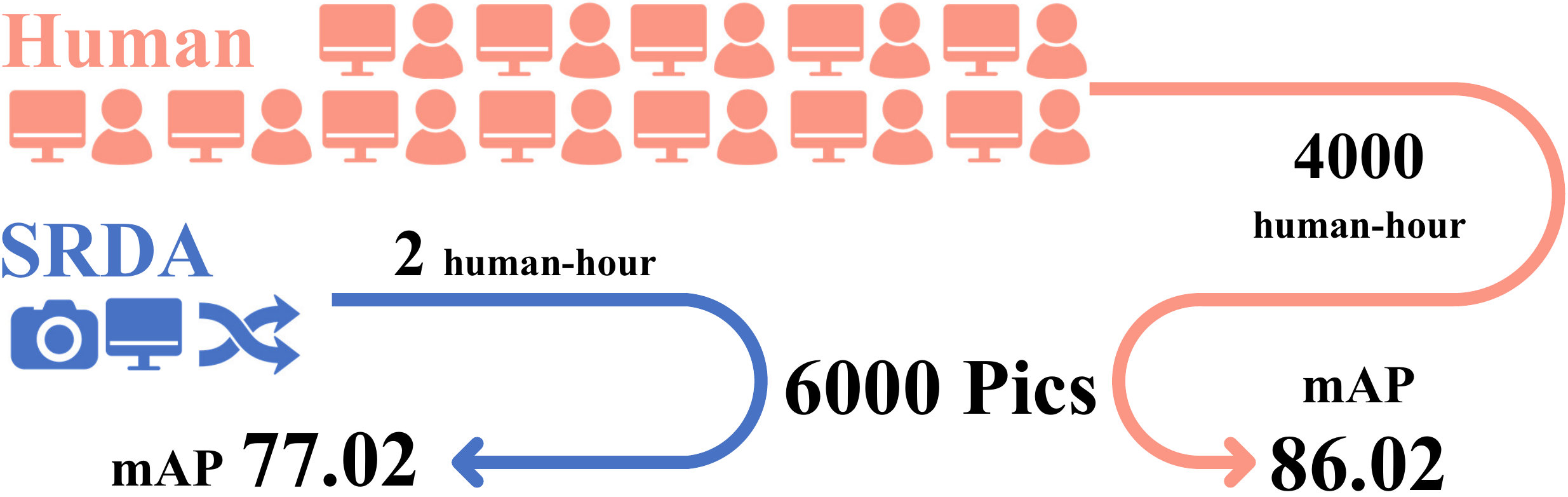}
	\end{center}
    \vspace{-0.4cm}
	\caption{Compared with human labeling (red), our pipeline (blue) can significantly reduce human labor cost by nearly 2000 folds and achieve reasonable accuracy in instance segmentation. 77.02 and 86.02 are average mAP@0.5 of 3 scenes.}\protect\label{Figure:HumanTime}
	\vspace{-0.5cm}
\end{figure}

We present a new pipeline that attempts to address these challenges. Our pipeline comprises three stages: \emph{scanning}, \emph{physics reasoning}, \emph{domain adaptation} ({\bf SRDA}) as shown in Fig. \ref{Figure:HumanTime}. At the first stage, new objects and environmental background from a certain scene are scanned into 3D models. Unlike other CG based methods that do simulation with existing model datasets, images synthesized by our pipeline can ensure realistic effect and well describe the targeting environment, since we use real-world scanned data. At the reasoning stage, we proposed a reasoning system to generate proper layout for each scene by fully considering physically and commonsense plausible. Physics engine is used to ensure physically plausible and commonsense plausible is checked by commonsense likelihood (CL) function. For example, ``a mouse on the mouse pad and they on the table'' would have a large output in CL function. In the last stage, we proposed a novel \emph{Geometry-guided GAN} (GeoGAN) framework. It integrates geometry information (segmentation as edge cue, surface normal, depth) which helps to generate more plausible images. In addition, it includes a new component \emph{Predictor} which can serve as a useful auxiliary supervision, and also a criterion to score the visual quality of images.

The major advantage of our pipeline is time-saving. Compared with conventional exhausting annotation, we can reduce labor cost by nearly 2000 folds, in the meantime, achieve decent accuracy, preserving 90\% performance. (See Fig. \ref{Figure:HumanTime}). The most time-consuming stage is scanning, which is easy to accomplish in most of indoor and some of outdoor scenarios.

Our pipeline can be widely adaptive to many scenarios. We choose three representative scenes, namely a shelf from a supermarket (for a self-service supermarket), a desk from an office (for home robot), a tote similar in Amazon Robotic Challenge\footnote{https://www.amazonrobotics.com/\#/roboticschallenge}. 
	
To the best of our knowledge, no current datasets consist of compact 3D object/scene models and real scene images with instance segmentation annotations. Hence, we build a dataset to prove the efficacy of our pipeline. This dataset have two parts, one for scanned object models (SOM dataset) and one for real scene images with instance level annotations (Instance-60K).
	
Our contributions have two folds: 
\begin{itemize}
	\item The main contribution is the novel three-stage SRDA pipeline. We added a reasoning system to the feasible layout building and proposed a new domain adaptation framework named GeoGAN. It is time-saving and the output images are close to real ones according to the evaluation experiment.
	\item To demonstrate the effectiveness, we build up a database which contains 3D models of common objects and corresponding scenes (SOM dataset) and scene images with instance level annotations (instance-60K).
\end{itemize}
	
We will first review some of the related concepts and works in Sec. \ref{relwork} and depict the whole pipeline from Sec. \ref{sec:scan} on. We describe the scanning process in Sec. \ref{sec:scan}, reasoning system in Sec. \ref{sec:sim}, and GAN-based domain adaptation in Sec. \ref{sec:da}. In Sec. \ref{sec:dataset}, we illustrate how Instance-60K dataset is built. Extensive evaluation experiments are carried out in Sec. \ref{sec:eval}. And finally, we discuss the limitation of our pipeline in Sec. \ref{sec:dis}.

\section{Related Works}
\label{relwork}
{\bf Instance Segmentation} Instance segmentation has become a hot topic in recent years. Dai et al. \protect\cite{MdNC} proposed a complex multiple-stage cascaded network that does detection, segmentation, and classification in sequence. Li et al. \protect\cite{YiLi2017} combined a segment proposal system and object detection system, simultaneously producing object classes, bounding boxes, and masks. Mask R-CNN \protect\cite{he2017mask} supports multiple tasks including instance segmentation, object detection, human pose estimation. Whereas exhausting labeling is required to guarantee a satisfactory performance, if we apply these methods to a new environment.

{\bf Generative Adversarial Networks} Since introduced by Goodfellow \protect\cite{goodfellow2014generative}, GAN-based methods have fruitful results in various fields, such as image generation \protect\cite{Radford2015}, image-to-image translation \protect\cite{CycleGAN2017}, 3D model generation \protect\cite{wu2016learning}, etc. The former paper on image-to-image translation inspired our work, it indicates GAN has the potential to bridge the gap between simulation domain and real domain.
	
{\bf Image-to-Image Translation} A general image-to-image translation framework was first introduced by Pix2Pix \protect\cite{Isola_2017_CVPR}, but it required a great amount of paired data. Chen \protect\cite{ChenK17aa} proposed a cascaded refinement network free of adversarial training, which gets high-resolution results, but still demands paired data. Taigman et al. \protect\cite{Taigman2016Unsupervised} proposed an unsupervised approach to learn cross-domain conversion, however it needs a pre-trained function to map samples from two domains into an intermediate representation. Dual learning \protect\cite{CycleGAN2017,yi2017dualgan,kim2017learning} is soon imported for unpaired image translation, but currently, dual learning methods encounter setbacks when camera viewpoint or object position varies. On the contrary to CycleGAN, Benaim et al. \protect\cite{Benaim2017One} learned one-side mapping.
Refining rendered image using GAN is also not unknown \protect\cite{sixt2016rendergan,Shrivastava2016Learning,Bousmalis2016Unsupervised}.
Our work is a complementary to these approaches, where we deal with more complex data and tasks. We will compare \protect\cite{Shrivastava2016Learning,Bousmalis2016Unsupervised} with our GeoGAN in Sec. \ref{sec:eval}. 
	
{\bf Synthetic Data for Training} Some researchers attempt to generate synthetic data for vision tasks such as viewpoint estimation \protect\cite{Su_2015_ICCV}, object detection \protect\cite{georgakis2017synthesizing}, semantic segmentation \protect\cite{RosCVPR16}. In \protect\cite{alhaija2017augmented}, Alhaija et al. addressed generation of instance segmentation training data for street scenes with technical effort in producing realistically rendered and positioned cars. However, they focus on street scenes and do not use an adversarial formulation.
	
{\bf Scene Generation by Computer Graphics} Scene generation by CG techniques is a well-studied area in the computer graphics community \protect\cite{Handa2016SceneNet,Mccormac2017SceneNet,Song2016Semantic,Fisher2012Example,Merrell2011Interactive}. These methods are capable of generating plausible layout of indoor or outdoor scene, but they have no intention to transfer the rendered images to real domain.

\section{Scanning Process}
\label{sec:scan}
In this section, we describe the scanning process.
Objects and scene backgrounds are scanned in two ways due to the scale issue.
	
We choose the multi-view environment (MVE) \protect\cite{fuhrmann2014mve} to perform dense reconstruction for objects, since it is image-based and thus requires only a RGB sensor. Objects are first videotaped, which can be easily done by most RGB sensors. In the experiment, we use an iPhone5s. The videos are sliced into images with multiple viewpoints, and fed into MVE to generate 3D models.
We can videotape multiple objects (at least 4) and generate corresponding models per time, which can alleviate the scalability issue when new objects are too many to scan one by one.
MVE is capable of generating dense meshes with a fine texture. As for the texture-less objects, we scan the object with hand holding, and the hand-object interaction can be a useful cue for reconstruction, as indicated in \protect\cite{tzionas20153d}.
	
For the environmental background, scenes without targeting objects were scanned by Intel RealSense R200 and reconstructed by ReconstructMe\footnote{http://reconstructme.net/}. We follow the official instruction to operate reconstruction.
	
Resolution for iPhone5s is 1920$\times$1080 and for R200 is 640$\times$480 at 60 FPS. Remaining settings are by default.

\begin{figure}[!ht]
	\begin{center}
		\includegraphics[width=11cm,height=4.8cm]{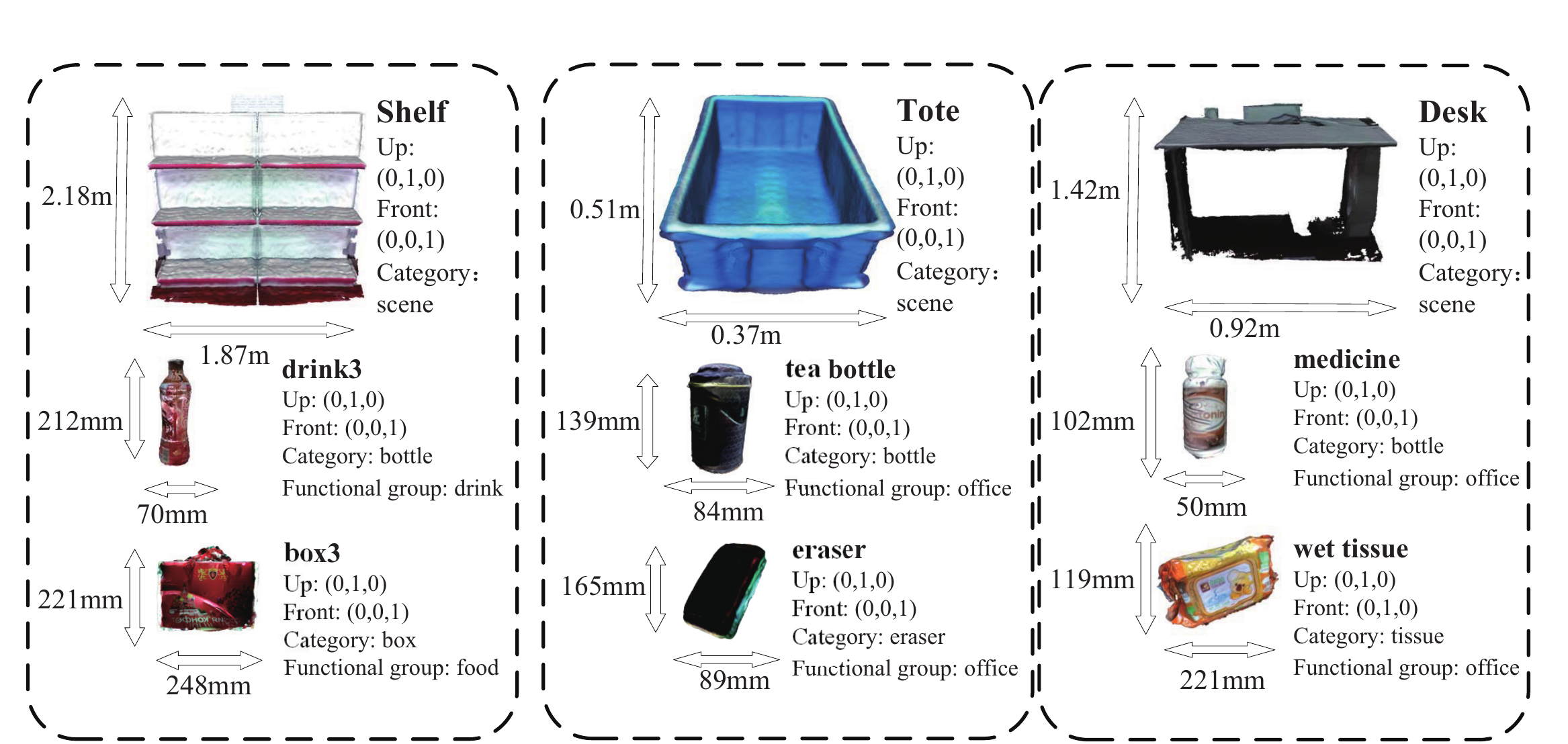}
	\end{center}
	\vspace{-0.5cm}
	\caption{Representative environmental backgrounds, object models, and corresponding label information.}
	\protect\label{fig:representative}
\end{figure}
	
\section{Layout Building With Reasoning}
\vspace{-0.8cm}
\label{sec:sim}
\begin{figure}[!ht]
	\begin{center}
		\includegraphics[width=10cm,height=4.2cm]{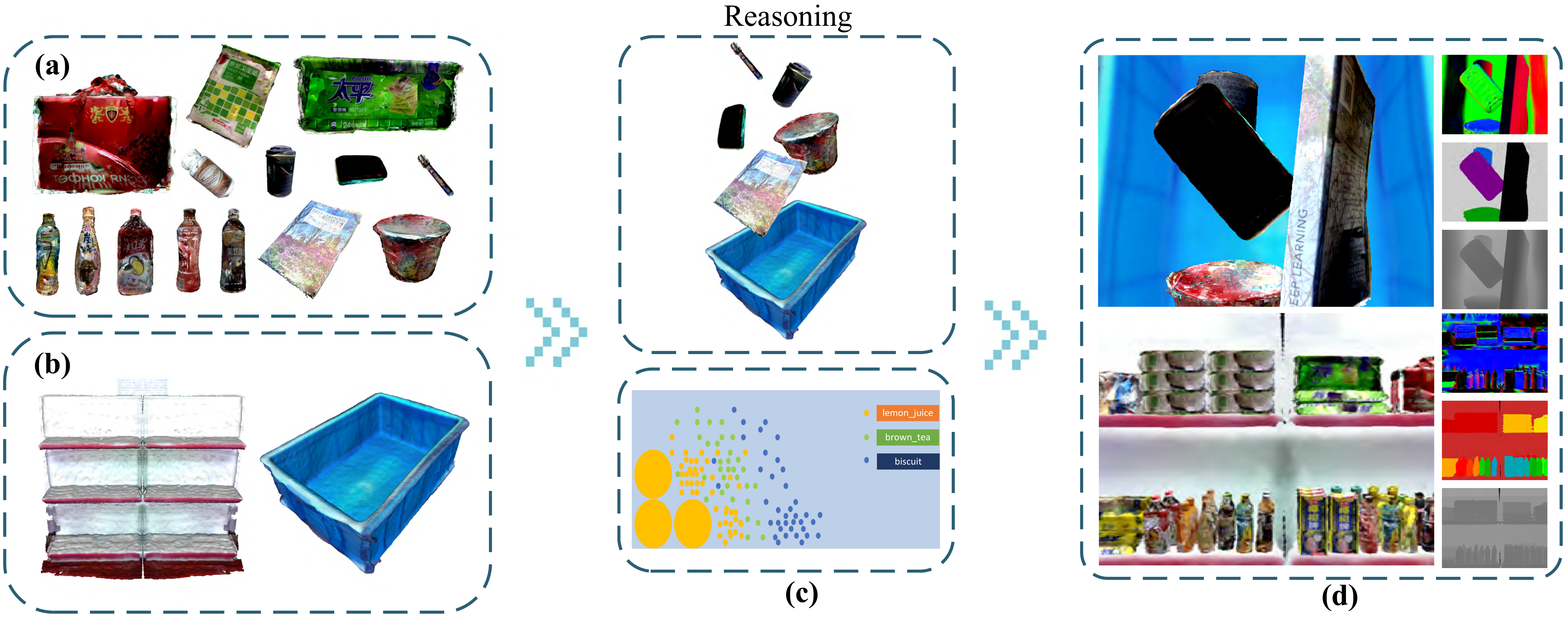}
	\end{center}
	\vspace{-0.5cm}
	\caption{The scanned objects (a) and background (b) are put into a rule-based reasoning system (c) to generate physically plausible layouts. The upper of (c) is the random scheme, while the bottom is the rule-based scheme. In the end, system output rough RGB images and corresponding annotations (d). }\protect\label{fig:renderPipe}
	\vspace{-0.5cm}
\end{figure}

\subsection{Scene Layout Building With Knowledge}
With 3D models of objects and environmental background at hand, we are ready to generate scenes by our reasoning system. A proper scene layout must obey physics laws and human conventions. To make scene physically plausible, we select an off-the-shelf physics engine, Project Chrono \protect\cite{Chrono2016}. However, it is not as direct to make object layout convincing, some commonsense knowledge should be incorporated. To produce a feasible layout, we need to make object pose and location reasonable. For example, a cup always has the pose of ``standing up'', but not ``lying down'', meanwhile, it is always on the table not the ground. This prior falls in common daily knowledge that can not be achieved by physics reasoning. Therefore, we present how to annotate the pose and location prior in what follows. 

\textbf{Pose Prior}: For each object, we show annotators its 3D model in 3D graphics environment, and ask annotators to draw all its possible poses that she/he can imagine. For each possible pose, the annotator should suggest a probability that this pose would happen. We record the the probability of $i^{th}$ object in pose $k$  as $D_p[k|i]$. We use interpolation to ensure most of pose has correponding probability value.

\textbf{Location Prior}: The same as pose prior, we show annotators the environmental background in 3D graphics environment, thus annotators label all its possible locations that an object may be placed. For each possible location, the annotator should suggest a probability that this object would be placed. We denoted the probability of $i^{th}$ object in location $k$  as $D_l[k|i]$. We use interpolation to make most of location has correponding probability value.

\textbf{Relationship Prior}: Some objects have strong co-occurrence prior. For example, mouse is always close to laptop. Given an object name list, we use language prior to select a set of object pair that have high co-occurrence probability, we call them as occurrence object pair (OOP). For each OOP,  annotator suggests a probability of occurrence of corresponding object pairs. For object $i^{th}$ and $j^{th}$, their probability of occurrence is denoted as $D_r[i,j]$ and a suggested distance (by annotators) is $H_r[i,j]$.

Note that the annotation maybe subjective, but we found that we only need a prior for layout generation guidance. Extensive experiments show that roughly subjective labeling is sufficient for producing satisfactory results. We will report the experiment details in supplementary file. 

\subsection{Layout Generation by Knowledge}
We generate layout by considering both physics laws and human conventions. First, we randomly generate a layout and check its physically plausible by Chrono. If it is not physically reasonable, we reject this layout. Second, we check its commonsense plausible by three priors above. In detail, all object pairs are extracted in layout scene. We denote $(\{c_1(i),c_2(i)\}$, $(\{p_1(i),p_2(i)\}$ and $(\{l_1(i),l_2(i)\}$ as category, pose and 3D location of $i^{th}$ extracted object pair in scene layout. The likelihood of pose is expressed as 
\begin{equation}\label{eq:pose_prior_1}
	K_p[i] = D_p[p_1(i)|c_1(i)] D_p[p_2(i)|c_2(i)].  
\end{equation}
The likelihood of location for $i^{th}$ object pair is written as,
\begin{equation}\label{eq:pose_prior_2}
	K_l[i] = D_l[l_1(i)|c_1(i)] D_l[l_2(i)|c_2(i)].  
\end{equation}
The likelihood of occurrence for $i^{th}$ object pair is presented as
\begin{equation}\label{eq:pose_prior_3}
	K_r[i] = \begin{cases}
    G_{\sigma}(|l_1(i)- l_2(i)| - D_r[c_1(i),c_2(j)]) & \text{if } H_r[i,j] > \gamma\\1,& \text{otherwise}.\end{cases}
\end{equation}
where $G_{\sigma}$ is a Gaussian function with parameter $\sigma$ ($\sigma=0.1$ in our paper). We compute occurrence prior in the case where the probability $H_r[i,j]$  is larger than a threshold $\gamma$ ($\gamma = 0.5$ in our paper).   

We denote commonsense likelihood function of a scene layout as 
\begin{equation}\label{eq:prior1}
	K =  \prod_i K_l[i] K_l[i] K_r[i] \propto \sum_i \log(K_l[i]) + \log(K_p[i]) + \log(K_r[i])  
\end{equation}
Thus, we can judge commonsense plausible by $K$. If $K$ is smaller than a threshold, we reject its corresponding layout. In this way, we can generate large quantities of layouts that is both physics and commonsense plausible.
 
\subsection{Annotation Cost} We annotate scanned model one by one. So, the annotation cost is linear scale with respect to scanned object model number $M$. Note that only a small set of object have strong object occurrence assumption (e.g. laptop and mouse). So, the complexity of object occurrence annotation is close to $O(M)$. We carry out experiment to find that 10 seconds is taken to label knowledge for a scanned object model in average, which is minor (one hour for hundreds of objects).  
    
\section{Domain Adaptation With Geometry-guided GAN}
\label{sec:da}
Now, we have collection of the rough (RGB) image $\lbrace I_i^{r} \rbrace^M_{i=1} \in \mathcal{I}^{r}$ and its corresponding ground truths, instance segmentation $\lbrace I_i^{s\text{-}gt} \rbrace^M_{i=1} \in \mathcal{I}^{s\text{-}gt}$, surface normal $\lbrace I_i^{n\text{-}gt} \rbrace^M_{i=1} \in \mathcal{I}^{n\text{-}gt}$, depth image $\lbrace I_i^{d\text{-}gt} \rbrace^M_{i=1} \in \mathcal{I}^{d\text{-}gt}$. Besides, the real image captured from targeting environment is denoted as $\lbrace I_j\rbrace^N_{j=1}$. $M,N$ are the sample sizes for rendered samples and real samples. With these data, we can embark on training GeoGAN.
	
\begin{figure}[!ht]
	\begin{center}
		\includegraphics[width=11cm,height=3.1cm]{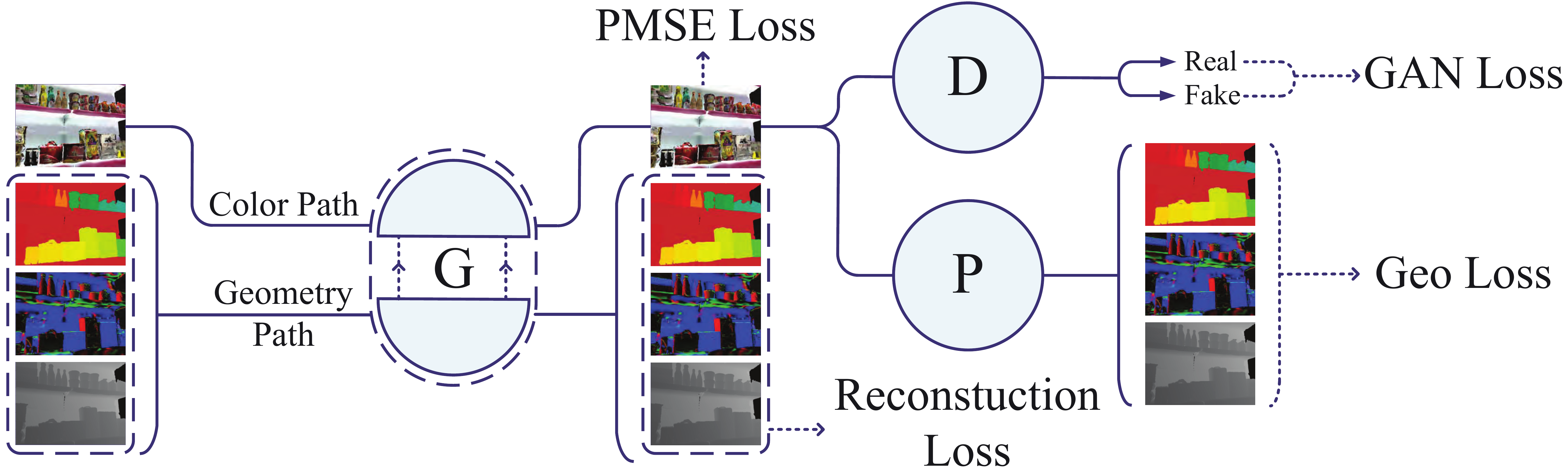}
	\end{center}
    \vspace{-0.2cm}
	\caption{The GDP structure consists of three components: a generator (G), a discriminator (D), and a predictor (P), along with four loss: LSGAN loss (GAN loss), Structure loss, Reconstruction loss (L1 loss), Geometry-guided loss (Geo loss).}\protect\label{fig:GDP}
    \vspace{-0.5cm}
\end{figure}

\subsection{Objective Function}
GeoGAN has a ``GDP" structure, as sketched in Fig. \ref{fig:GDP}, which comprises a generator (G), a discriminator (D) and a predictor (P) which serves as a geometry prior guidance. Such structure leads to the design of the objective function, which consists of four loss functions that will be presented in what follows.
	
{\bf \emph{LSGAN Loss}} We adopt a least-square generative adversarial objective (LSGAN)\protect\cite{Mao2016Least} to help G and D training stable. The LSGAN adversarial loss can be written as
\begin{equation}\label{eq:GAN}
\mathcal{L}_{GAN}(G,D)=\mathbb{E}_{y\sim p_{data}(y)}[(D(y)-1)^2 ]+\mathbb{E}_{x\sim p_{data}(x)}[(D(G(x)))^2],
\end{equation}
$x$ and $y$ stand for a sample from the rough image and the real image domain respectively.
	
We denote the output of the generator with parameter $\Phi_G$ for $i^{th}$ rough image as $I^{*}_{i}$, i.e. $I^{*}_{i}\triangleq G(I^r_i|\Phi_G)$
	
{\bf \emph{Structure Loss}} A structure loss is introduced to ensure $I^{*}_i$ maintains the original structure of $I^r_i$. A Pairwise Mean Square Error (PMSE) loss is imported from \protect\cite{Eigen2014Depth}, expressed as:
\begin{equation}\label{eq:PMSE}	
\mathcal{L}_{PMSE}(G)=\frac{1}{N}\sum_{i}(I_i^{r}-I^{*}_{i})^2-\frac{1}{N^2}(\sum_{i,j}(I_i^{r}-I^{*}_{i}))^2.
\end{equation}
	
{\bf \emph{Reconstruction Loss}} To ensure the geometry information successfully encoded in the network. We also use $\ell^1$ as a reconstruction loss for the geometric images.
\begin{equation}\label{eq:recon}
\begin{aligned}
\mathcal{L}_{rec}(G)=||[I^r,I^s,I^n,I^d|\Phi_G]_{rec},[I^r,I^s,I^n,I^d]||_1
\end{aligned}
\end{equation}
	
{\bf \emph{Geometry-guided Loss}} Given an excellent geometer predictor, a high-quality image should be able to produce desirable instance segmentation, depth map and normal map. It is a useful criterion that judges whether $I^{*}_{i}$ is qualified or not. An unqualified image (with artifacts, distorted structure) will induce large geometry-guided loss (Geo Loss).  
	
To achieve this goal, we pretrained the predictor with following formula:
\begin{equation}\label{eq:stm}
	[I^s, I^n, I^d] = P(I |\Phi_P),
\end{equation}
	
It means given an input image $I$, with the parameter $\Phi_P$, the predictor can output instance segmentation $I^s$, normal map $I^n$ and depth map $I^d$ respectively.
In the first few iterations, the predictor is pretrained with the rough image, that is, $I=I^r$. When the generator starts to produce reasonable results, $\Phi_P$ can be updated with $I=I^*$. And then, the predictor is ready to supervise the generator, and $\Phi_G$ will be updated as follow:
	
\begin{equation}\label{eq:geoloss}
\begin{aligned}
	\mathcal{L}_{Geo}(G,P) = ||P(I^{*}_i|\Phi_P) , [I^{s\text{-}gt}_i, I^{n\text{-}gt}_i, I^{d\text{-}gt}_i]||^2_2.
\end{aligned}
\end{equation}
In this equation, $\Phi_P$ is not updated, and it is a $\ell^2$ loss.
	
{\bf \emph{Overall Objective Function}}
In sum, our objective function can be expressed as:
\begin{equation}\label{eq:overall}
\begin{aligned}
	&\min_{\Phi_G}\max_{\Phi_D}   \lambda_1 \mathcal{L}_{GAN}(G,D) +\lambda_2 \mathcal{L}_{PMSE}(G)+\lambda_3 \mathcal{L}_{rec}(G)+\lambda_4 \mathcal{L}_{Geo}(G, P), \\
	&\min_{\Phi_P} \mathcal{L}_{Geo}(G, P).\\
\end{aligned}	
\end{equation}
	
\begin{figure}[!ht]
	\begin{center}
		\includegraphics[width=7cm,height=2cm]{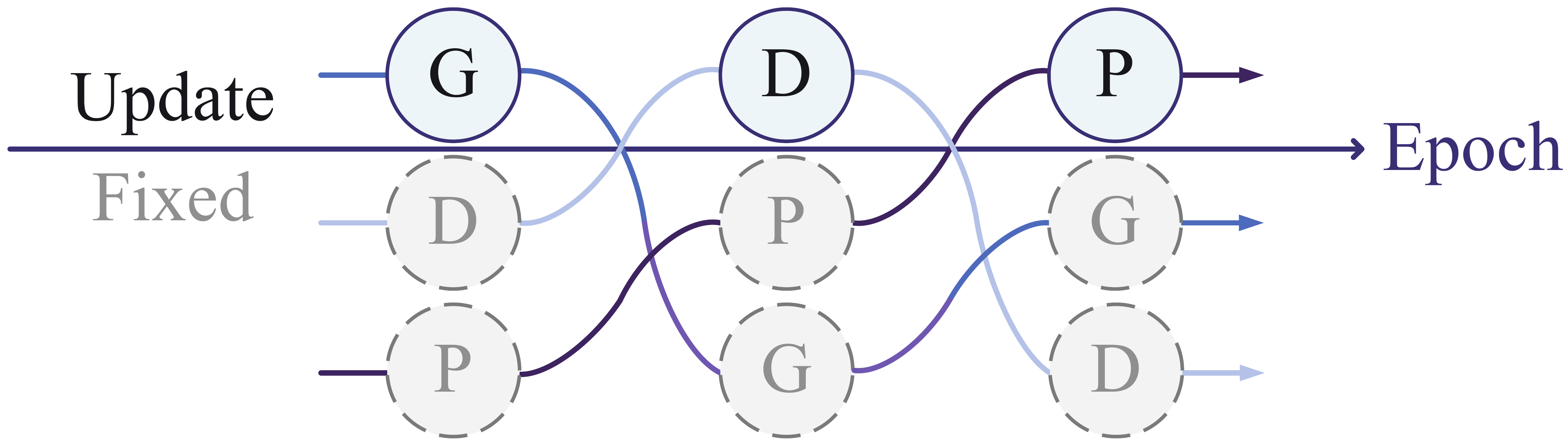}
	\end{center}
    \vspace{-0.5cm}
	\caption{Iterative optimization framework. As the epoch goes, G, D and P are updated as presented. While one component is updating, the other two are fixed.}\protect\label{fig:iterative}
\end{figure}

This objective function reveals the iterative nature of our GeoGAN framework, as sketched in Fig. \ref{fig:iterative}.

\subsection{Implementation}
	
We will dive into details of how to implement and train our model. 
	
{\bf Dual Path Generator (G)}
Our generator has dual forward data paths (color path and geometry path), which help to integrate the color and geometry information. For color path, input rough image will firstly pass three convolutional layers, and then downsample to $64\times 64$ and pass 6 resnet blocks \protect\cite{He2016Deep}. After that, output feature maps are upsampled to $256\times 256$ with bilinear upsampling. During upsampling, color information path will concatenate feature maps from geometry information path.
	
Geometry information are firstly convolutioned to feature maps and concatenated together, resulting in a 3-dimensional $256\times 256$ feature map before passing to geometry path described below. After the last layer, we split the output of the last layer into three parts, and produce three reconstruction images for three kinds of geometric images.
	
Let $3n64s1ReLU$ denote $3\times 3$-Convolution-InstanceNorm-ReLU layer with 64 filters and stride 1. $Rk$ denotes a residual block that contains two $3\times 3$ convolutional layers with the same number of filters on both. $upk$ denotes a bilinear upsampling layer followed with a $3 \times 3$ Convolution-InstanceNorm-ReLU layer with k filters and stride 1.
	
The generator architecture is:
	
color path: 7n3s1ReLU-3n64s2ReLU-3n128s2ReLU-R256-R256-R256-R256\\-R256-R256-up512-up256
	
geometry path: 7n3s1ReLU-3n64s2ReLU-3n128s2ReLU-R256-R256-R256\\-R256-R256-R256-up256-up128
	
{\bf Markovian Discriminator (D)}
The discriminator is a typical PatchGAN or Markovian discriminator described in \protect\cite{Li2016Precomputed,Ledig2016Photo,Isola_2017_CVPR}. We also found 70$\times$70 is a proper receptive field size, hence the architecture is exactly like \protect\cite{Isola_2017_CVPR}.
	
{\bf Geometry Predictor (P)}
FCN-like networks\protect\cite{Long2015Fully} or UNet\protect\cite{Ronneberger2015U} are good candidates for the geometry predictor. In implementation, we choose a UNet architecture. $downk$ denotes a $3 \times 3$ Convolution-InstanceNorm-LeakyReLU layer with k filters and stride 2, the slope of leaky ReLU is 0.2. $upk$ denotes a bilinear upsampling layer followed with a $3 \times 3$ Convolution-InstanceNorm-ReLU layer with k filters and stride 1. $k$ in $upk$ is 2 times larger than that in $downk$, since a skip connection between corresponding layers. After the last layer, feature maps are split into three parts and convolution to a three dimension layer separately, activated by tanh function.

The predictor architecture is:
down64-down128-down256-down512-down512-down512-up1024-up1024-up1024-up512-up256-up128

{\bf Training Details} Adam optimizer\protect\cite{Kingma2014Adam} is used for all three ``GDP" components, with batch size of 1. G, D and P are trained from scratch. We firstly trained geometry predictor with 5 epochs to get a good initialization, then began the iterative procedures. In the iterative procedures, learning rate for the first 100 epochs are 0.0002 and linearly decay to zero in the next 100 epochs. All training images are of size $256\times 256$.
	
All models are trained with $\lambda_1=2$, $\lambda_2=5$, $\lambda_3=10$, $\lambda_4=3$ in Eq. \ref{eq:overall}. The generator is trained twice before the discriminator updates once.

\section{Instance-60K Building Process}
\label{sec:dataset}	
As we found no existing Instance segmentation datasets \protect\cite{Cordts2016Cityscapes,Lin2014Microsoft,pascal-voc-2012} can benchmark our task, we have to build a new dataset to benchmark our method. 
	
Instance-60K is an ongoing effort to annotate instance segmentation for scenes can be scanned. Currently it contains three representative scenes, namely supermarket shelf, office desk and tote. These three scenes are chosen since they potentially benefit real-world applications in the future. Supermarket cases are well-suited to self-service supermarkets like Amazon Go\footnote{https://www.amazon.com/b?node=16008589011}. Home robots will always meet the scene of an office desk. The tote is in the same setting as Amazon Robotic Challenge. 
	
\begin{figure}[!ht]
	\begin{center}
		\includegraphics[width=12cm,height=3cm]{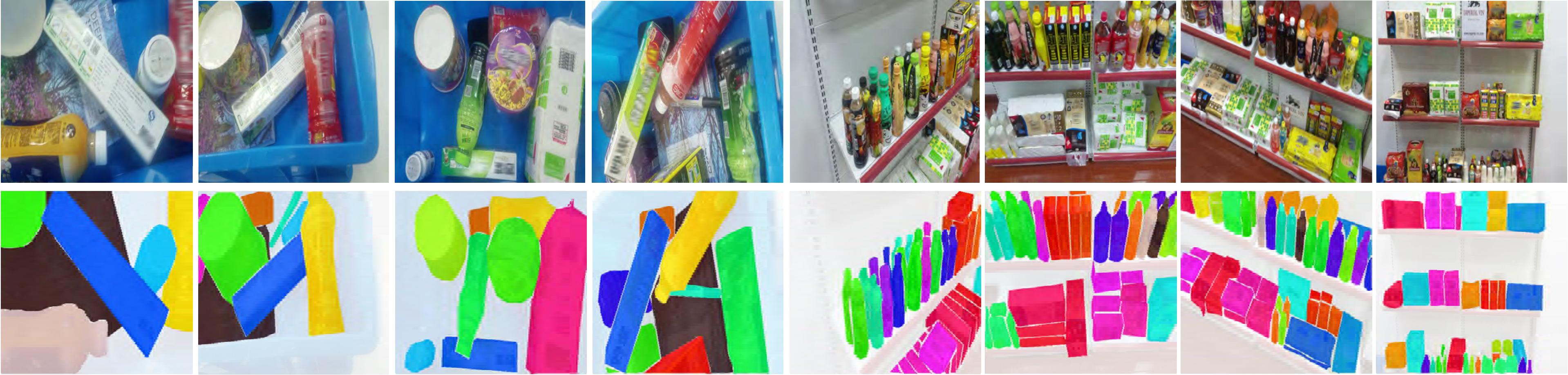}
	\end{center}	
    \vspace{-0.5cm}
	\caption{Representative images and manual annotations in the Instance-60K dataset.}
	\protect\label{fig:hard}
    \vspace{-0.5cm}
\end{figure}
	
To note that our pipeline does not restrict to these three scenes, technically any scenes can be simulated are suitable to our pipeline.
	
Shelf scene has objects of 30 categories, which items such as soft drinks, biscuits, and tissues. 15 categories for desk scene and tote scene. All are common objects in the corresponding scenes. Objects and scenes are scanned for building SOM dataset as described in section \ref{sec:scan}.
	
For instance-60K dataset, these objects are placed in corresponding scenes and then videotaped by iPhone5s under various viewpoints. We arranged 10 layouts for the shelf, and over 100 layouts for desk and tote. Videos are then sliced into 6000 images in total, 2000 for each scene. The number of labeled instance is 60894, that is the reason why we call it instance-60K. We have average 966 instances per category. This scale is about three times larger than PASCAL VOC \protect\cite{pascal-voc-2012} level (346 instances per category), so it is qualified to benchmark this problem. Again, we found instance segmentation annotation is laborious, it took more than 4000 man-hours on building this dataset. Some representative real images and annotation are shown in Fig. \ref{fig:hard}. As we can see, annotating them is time-consuming.
	
\section{Evaluation}
\label{sec:eval}
In this section, we evaluate our generated instance segmentation samples quantitatively and qualitatively.
	
\subsection{Evaluation on Instance-60K}
\vspace{-0.5cm}
\begin{table}[!ht]
	\begin{center}
		\begin{tabular}{C{2.22cm}C{1.22cm}C{1.26cm}C{1.26cm}C{1.26cm}}
			\toprule
			\multicolumn{3}{c}{mAP} & 0.5 & 0.7 \\ \hline
			\multirow{12}{*}{Mask R-CNN}  & \multirow{4}{*}{shelf} & real & 79.75 & 67.02 \\ \cline{3-5}
				&  & rough & 18.10 & 10.53 \\
				&  & fake & 49.11 & 37.56 \\
				&  & fake$_{plus}$ & {\bf 66.31} & {\bf 47.25} \\ \cline{2-5}
				& \multirow{4}{*}{desk} & real & 88.24 & 73.75 \\ \cline{3-5}
				&  & rough & 43.81 & 35.14 \\
				&  & fake & 57.07 & 45.44 \\
				&  & fake$_{plus}$ & {\bf 82.07} & {\bf 71.82} \\ \cline{2-5}
				& \multirow{4}{*}{tote} & real & 90.06 & 85.10 \\ \cline{3-5}
				&  & rough & 28.67 & 16.87 \\
				&  & fake & 61.40 & 50.13\\
				&  & fake$_{plus}$ & {\bf 82.69} & {\bf 76.84} \\ \bottomrule
		\end{tabular}
	\end{center}
	\caption{mAP results on real, rough, fake, fake$_{plus}$ models of different scenes with Mask R-CNN.}\protect\label{table:segRes}
    \vspace{-1cm}
\end{table}
	
\begin{figure}[!ht]
	\begin{center}
		\includegraphics[width=10cm,height=5cm]{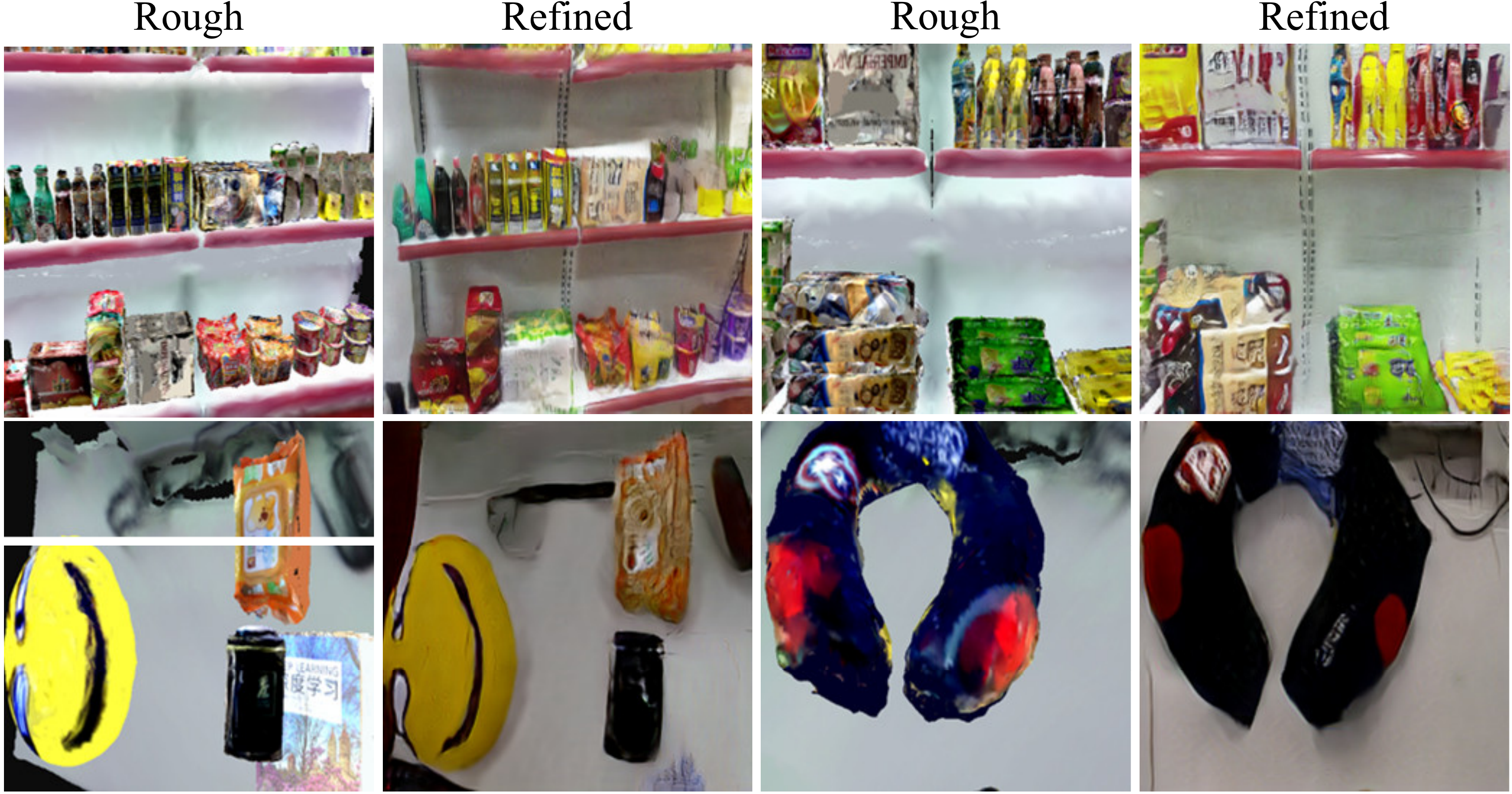}
	\end{center}
	\caption{Refinement of GAN. Refined column is the result of GeoGAN and rough column is the rendered image. Apparent improvement on lighting conditions and texture can be observed.}\protect\label{fig:refinePic}
    \vspace{-0.5cm}
\end{figure}
    
We employed instance segmentation tasks to evaluate on generated samples. To prove that the proposed pipeline generally works, we will report results using Mask R-CNN \protect\cite{he2017mask}.
	
We train segmentation model on resulting images produced by our GeoGAN. The trained model is denoted as ``fake-model''. Likewise, model trained on rough images is denoted as ``rough-model''. One question we should ask is that how about ``fake-model'' compare to models train on real images. To answer this question, we train segmentation models on training set of instance-60K dataset, which is denoted as ``real-model''. It is pre-trained on COCO dataset \protect\cite{Lin2014Microsoft}. 
	
Training procedures on real images strictly follow the procedures mentioned in \protect\cite{he2017mask}. We find the learning rate for real images is not workable to rough and GAN generated images, so we lower the learning rate and make it decay earlier.
	
All models are trained with 4500 images, though we can generate endless training sample for ``rough-model'' and ``fake-model'', since ``real-model'' only can train on 4500 images in the training set of instance-60K dataset. Finally, all models are evaluated on testing set of instance-60K dataset.

\begin{figure}[!ht]
	\begin{center}
		\includegraphics[width=11cm,height=4cm]{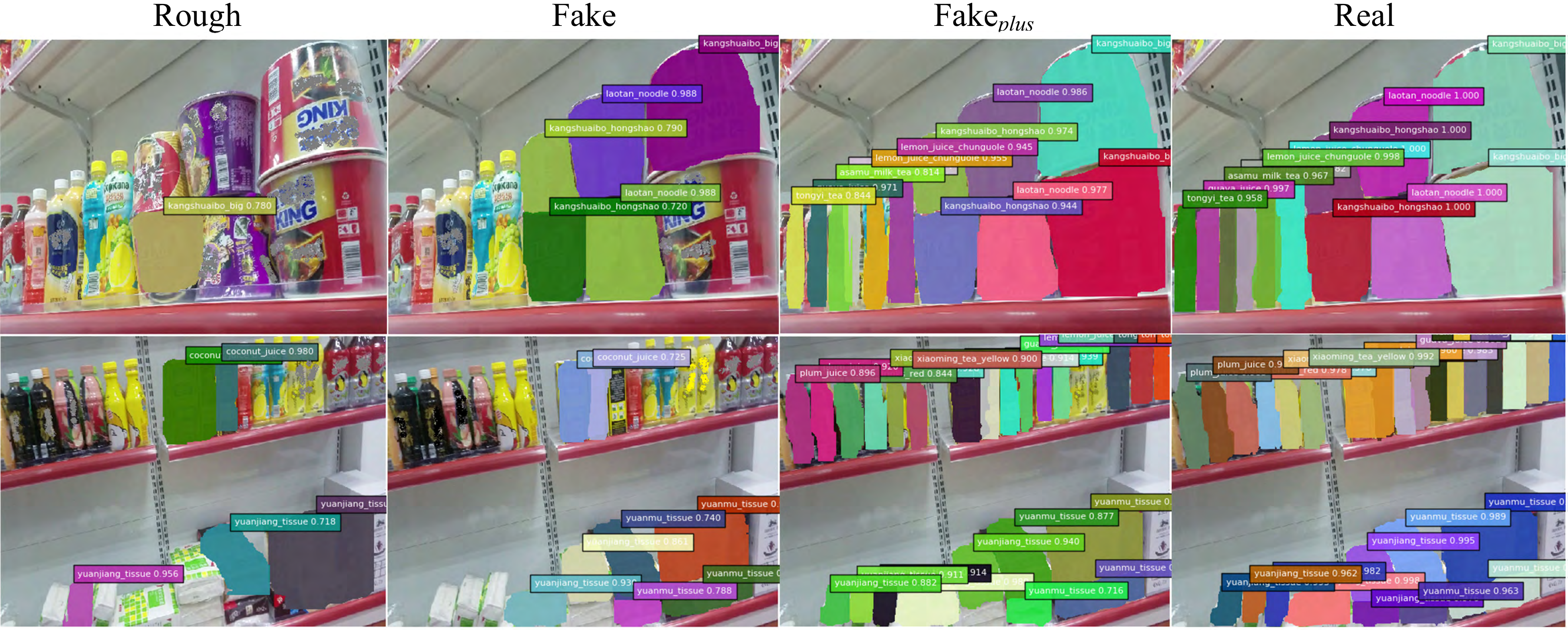}
	\end{center}
	\caption{Qualitative results visualization of rough, fake, fake$_{plus}$ and real model respectively.}
    \vspace{-0.5cm}
\end{figure}
	
Experiment results shown in Tab. \ref{table:segRes}. Overall mAP of the rough image is generally low, while ``fake-model'' significantly outperformed it. Noticeably, it still has a clear gap between ``fake-model" results and real one, though the gap has been bridged a lot.
	
Naturally, we would like to know how many refined training images is sufficient to achieve comparable results with ``real-model". Hence, we conducted experiments on 15000 GAN generated images, and named model as ``fake$_{plus}$-model". As we can see from Tab. \ref{table:segRes}, ``fake$_{plus}$" and ``real" is really close. We try to augment more training samples to ``fake$_{plus}$-model", but, the improvement is marginal. In this way, our synthetic ``images + annotation" is comparable with ``real image + human annotation" for instance segmentation.
	
The results for real-model may imply that our instance-60K is not that difficult for Mask R-CNN. Extension of the dataset is on-going. However, it is undeniable that the dataset is capable of proving the ability of GeoGAN.
	
In contrast to exhausting annotation using over 1000 human-hours per scene, our pipeline takes 0.7 human-hours per scene. Admittedly, the results suffer from performance loss, but save the whole task 3-order of human-hours.
	
\subsection{Comparison With Other Domain Adaptation Framework}
\begin{figure}[!ht]
	\begin{center}
		\includegraphics[width=8cm,height=4cm]{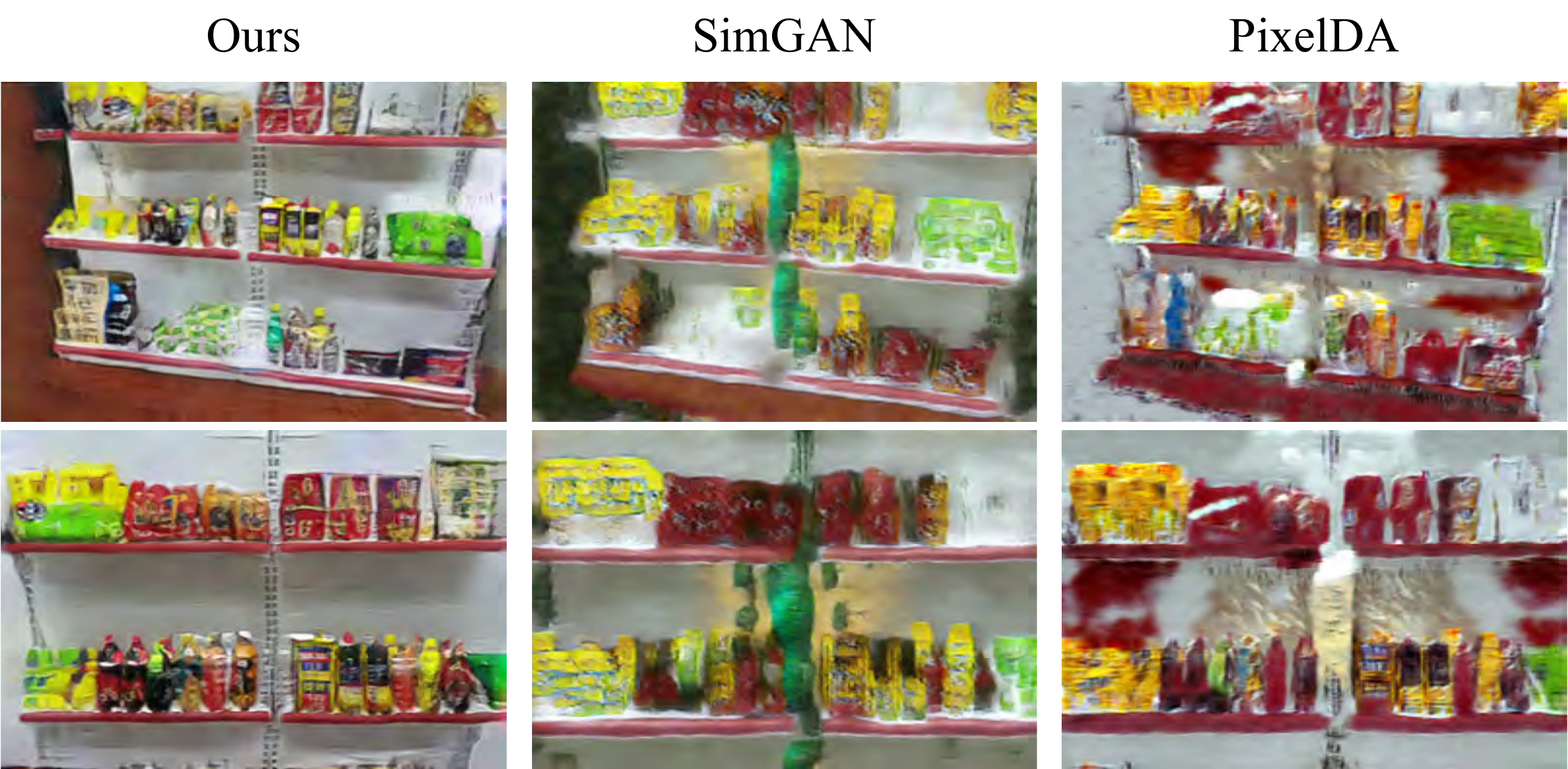}
	\end{center}
    \vspace{-0.5cm}
	\caption{Qualitative comparison of our pipeline and \protect\cite{Bousmalis2016Unsupervised}, \protect\cite{Shrivastava2016Learning}. The background of generated images from \protect\cite{Bousmalis2016Unsupervised} are damaged since they use a masked-PMSE loss.}\protect\label{fig:comparison}
    \vspace{-1.3cm}
\end{figure}

Previous domain adaptation framework focus on different tasks, such as gaze and hand pose estimation \protect\cite{Shrivastava2016Learning}, object classification and pose estimation \protect\cite{Bousmalis2016Unsupervised}. To the best of our knowledge, we are the first to propose a GAN-based framework to do instance segmentation. Comparison with each other is indirect. We reproduced the work of \protect\cite{Shrivastava2016Learning}  and \protect\cite{Bousmalis2016Unsupervised}. For \protect\cite{Bousmalis2016Unsupervised}, we substituted the task component with our P. The experiments are conducted on the scenes same in the paper. Results are shown in Fig.\ref{fig:comparison} and Tab.\ref{table:comparision}.
\begin{table}[!ht]
	\begin{center}
		\begin{tabular}{C{2.8cm}C{1.cm}C{2.36cm}C{1.0cm}C{1.0cm}}
			\toprule
				\multicolumn{3}{c}{mAP} & 0.5 & 0.7 \\ \hline
				\multirow{9}{*}{Mask R-CNN} & \multirow{3}{*}{shelf} & fake$_{plus, ours}$ & 66.31 & 47.25 \\ \cline{3-5}
				& & fake$_{plus, [25]}$ & 31.46 & 20.88 \\ \cline{3-5}
				&  & fake$_{plus, [13]}$ & 56.16 & 36.04 \\ \cline{2-5}
				& \multirow{3}{*}{desk}  & fake$_{plus, ours}$ &82.07 & 71.82 \\ \cline{3-5}
				& & fake$_{plus, [25]}$ & 44.33 & 29.93 \\ \cline{3-5}
				&  & fake$_{plus, [13]}$ & 69.54 & 57.27 \\ \cline{2-5}
				& \multirow{3}{*}{tote}  & fake$_{plus, ours}$ & 82.69 & 76.84 \\ \cline{3-5}
				& & fake$_{plus, [25]}$ & 42.50 & 33.61 \\ \cline{3-5}
				&  & fake$_{plus, [13]}$ & 70.73 & 62.68\\ \bottomrule
		\end{tabular}
	\end{center}
	\caption{Quantitative comparison of our pipeline and \protect\cite{Bousmalis2016Unsupervised}, \protect\cite{Shrivastava2016Learning}.}\protect\label{table:comparision}
    \vspace{-1.5cm}
\end{table}
	
\subsection{Ablation Study}
Ablation study is carried out by removing geometry-guided loss and structure loss separately. We applied Mask R-CNN to train the segmentation models on resulting images from GeoGAN without geometry-guided loss (denoted as ``fake$_{plus,\text{w/o-} geo}$-model") or structure loss (denoted as ``fake$_{plus, \text{w/o-}pmse}$-model"). As we can see, it suffers a significant performance loss when removing geometry-guided loss or structure loss. Besides, we also need to prove the necessity of reasoning system. After removing reasoning system, resulting in unrealistic images and performance loss. Results are shown in Tab. \ref{table:ablation}.
\begin{table}[!ht]
	\begin{center}
		\begin{tabular}{C{2.22cm}C{1.22cm}C{2.56cm}C{1.26cm}C{1.26cm}}
			\toprule
			\multicolumn{3}{c}{mAP} & 0.5 & 0.7 \\ \hline
			\multirow{12}{*}{Mask R-CNN} & \multirow{4}{*}{shelf} & fake$_{plus}$ & 66.31 & 47.25 \\ \cline{3-5}
			&  & fake$_{plus,\text{w/o-} geo}$ & 48.52 & 31.17 \\
			&  & fake$_{plus, \text{w/o-}pmse}$ & 27.33 & 19.24 \\
			&  & fake$_{plus, \text{w/o-}reason}$ & 15.21 & 8.44 \\ \cline{2-5}
			& \multirow{4}{*}{desk}  & fake$_{plus}$ & 82.07 & 71.82 \\ \cline{3-5}
			&  & fake$_{plus,\text{w/o-} geo}$ & 63.99 & 55.23 \\
			&  & fake$_{plus, \text{w/o-}pmse}$ & 45.05 & 34.51 \\
			&  & fake$_{plus, \text{w/o-}reason}$ & 18.36 & 9.71 \\ \cline{2-5}
			& \multirow{4}{*}{tote}  & fake$_{plus}$ & 82.69 & 76.84 \\ \cline{3-5}
			&  & fake$_{plus,\text{w/o-} geo}$ & 64.22 & 53.31\\
			&  & fake$_{plus, \text{w/o-}pmse}$ & 46.44 & 35.62 \\
			&  & fake$_{plus, \text{w/o-}reason}$ & 20.05 & 12.43\\ \bottomrule
		\end{tabular}
	\end{center}
	\caption{mAP results of ablation study on Mask R-CNN.}\protect\label{table:ablation}
    \vspace{-0.5cm}
\end{table}
\begin{figure}[!ht]
	\begin{center}
		\includegraphics[width=12cm,height=5.3cm]{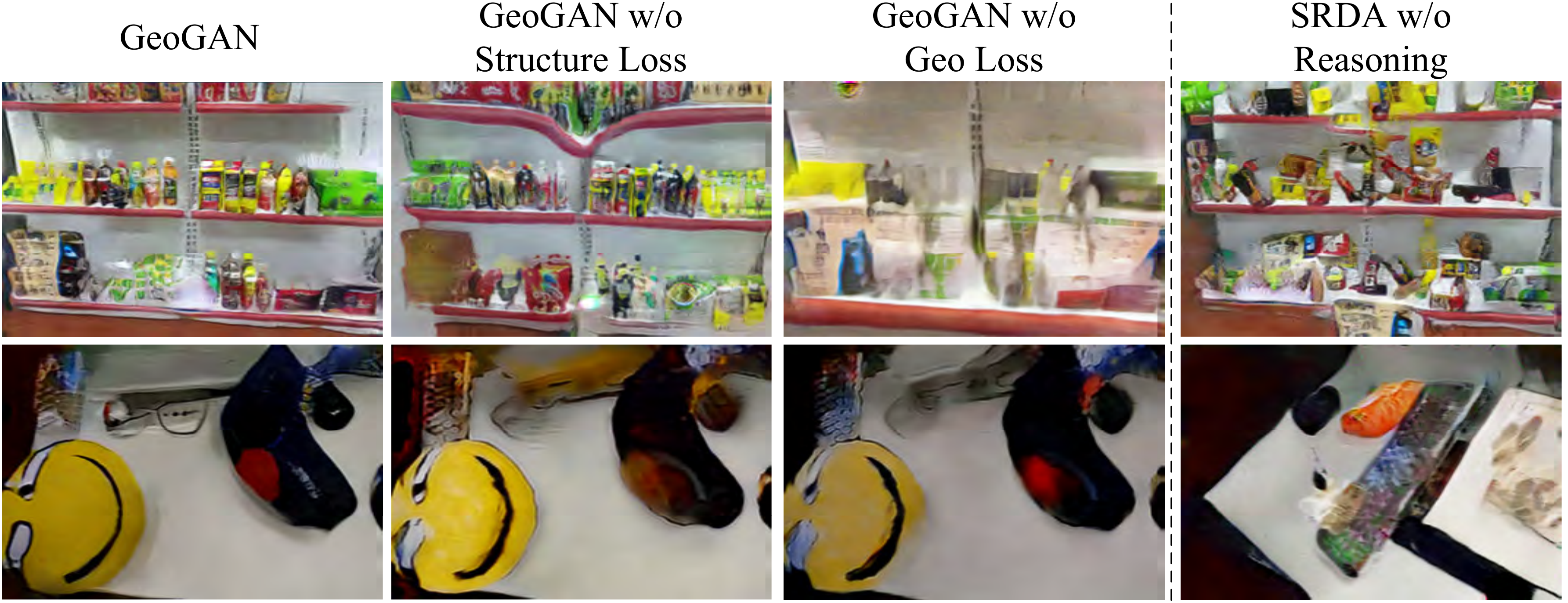}
	\end{center}
    \vspace{-0.5cm}
	\caption{Samples to illustrate the efficacy of structure loss, geometry-guided loss in GeoGAN and reasoning system in our pipeline.}\protect\label{fig:ablation}
\end{figure}
\vspace{-1.2cm}
	
\section{Limitations and Future Work}
\label{sec:dis}
If the environmental background changes dynamically, we should scan a large number of environmental backgrounds to cover this variance and take much effort. Due to the limitations of the physics engine, it is hard to handle highly non-rigid objects such as a towel. For another limitation, our method does not consider illumination effects in rendering, since it is much more complicated. GeoGAN that transfers illumination conditions of the real image may partially address this problem, but it is still imperfect. In addition, the size of our benchmark dataset is relatively small in comparison with COCO. Future work is necessary to address these limitations.

\bibliographystyle{splncs}
\bibliography{srda}

\section{Appendix}
\subsection{Extended Ablative Study}
We add experiments to address how each geometry information in the geometry path affects the results. As shown in table \ref{table:rebuttal}, model name with "w/o" means remove corresponding geometric information, while name without "w/o" means only corresponding geometric information is used for geometry path. As we can see, geometry path have some contributions for the final results, but not as much as geometry loss.

\vspace{-0.4cm}
\begin{table}[!ht]
    \begin{center}
        \begin{tabular}{C{2.22cm}C{1.22cm}C{3.2cm}C{1.26cm}C{1.26cm}}
            \toprule
            \multicolumn{3}{c}{mAP} & 0.5 & 0.7 \\ \hline
            \multirow{8}{*}{Mask R-CNN}  & \multirow{8}{*}{shelf} & fake$_{plus}$ & 66.31 & 47.25 \\ \cline{3-5}
            &  & fake$_{plus, w/o\text{-}geopath,geo}$ & 43.68 & 27.40\\
            &  & fake$_{plus, depth}$ & 57.80 & 39.52 \\
            &  & fake$_{plus, normal}$ & 55.45 & 36.18 \\
            &  & fake$_{plus, seg}$ & 55.23 & 37.89 \\
            &  & fake$_{plus, w/o\text{-}depth}$ & 61.12 & 42.09 \\
            &  & fake$_{plus, w/o\text{-}normal}$ & 62.87 & 46.26 \\
            &  & fake$_{plus, w/o\text{-}seg}$ & 58.05 & 41.77 \\ \bottomrule
        \end{tabular}
    \end{center}
    \caption{extended ablative experiments on geometric information.}
    \protect\label{table:rebuttal}
    \vspace{-1cm}
\end{table}

\subsection{Knowledge Acquiring With Many Or One}
\begin{figure}[!ht]
	\begin{center}
		\includegraphics[width=12cm,height=8.5cm]{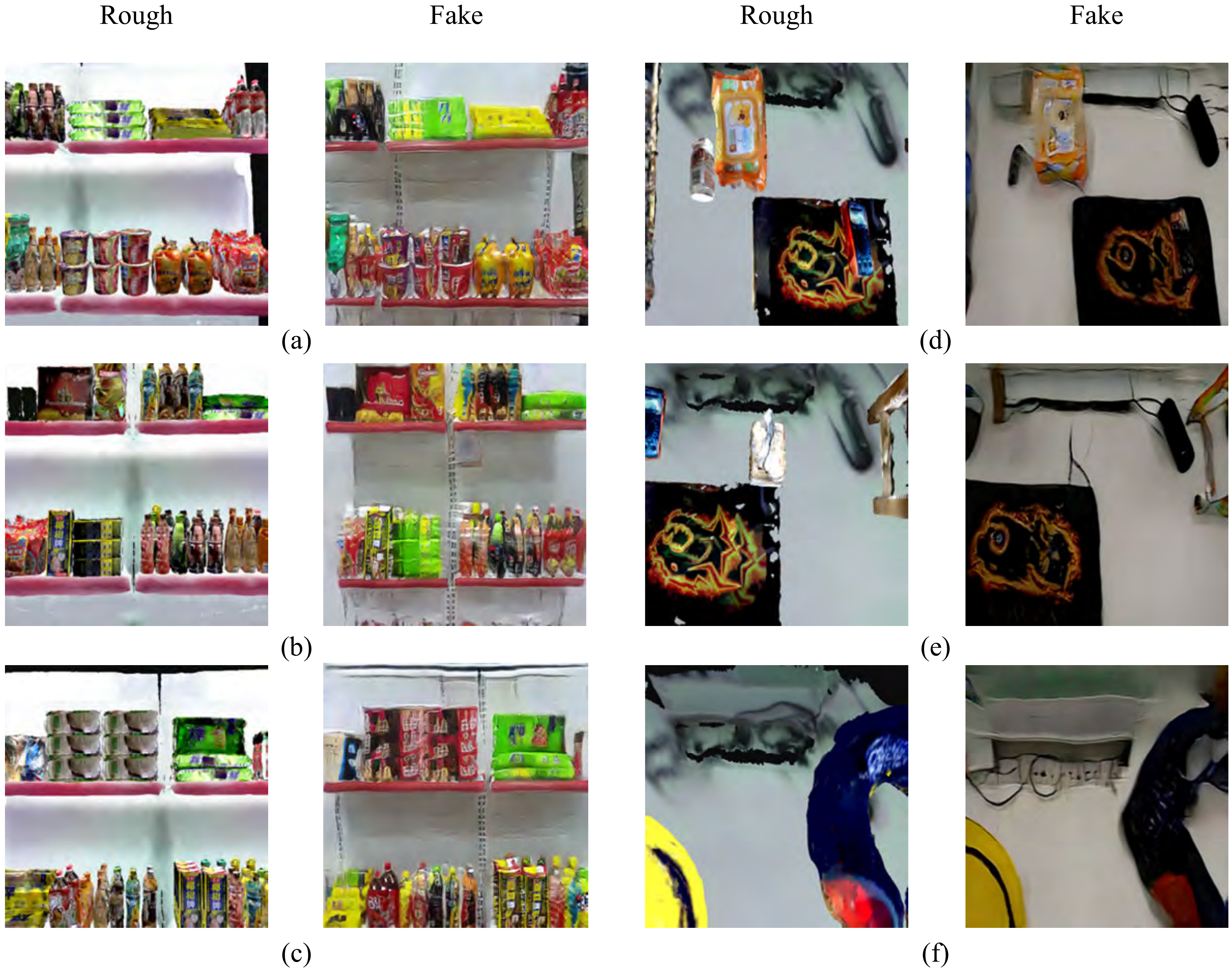}
	\end{center}
	\caption{Sample rough images generated from layouts which are synthesized by different people's annotation, and associated fake images. Annotator ID: (a)1, (b)3, (c)7, (d)8, (e)11, (f)18.}\protect\label{fig:know}
\end{figure}

In the experiment, knowledge (pose prior, location prior and relationship prior) annotated on objects can be acquired with ease. One or two people are more than enough to handle the workload. Nonetheless, annotation in this way seems subjective at the first glance. What if one annotator thinks object A should stands upright in scene B, which the other thinks otherwise. We admit the cognitive bias exists as pointed out by \protect\cite{greene2016estimations}. However, to our surprise, experiments show that such bias does not have that significant influence on the domain adaption results (See Fig. \ref{fig:know}) and segmentation results (See Tab. \ref{table:know}).

\begin{longtable}{C{2.8cm}C{1.cm}C{2.36cm}C{1.0cm}C{1.0cm}}
        	\toprule
			\multicolumn{3}{c}{mAP} & 0.5 & 0.7 \\ \hline
			\multirow{3}{*}{Annotator 1} & shelf & fake$_{plus}$ & 66.31 & 47.25 \\ \cline{2-5}
			& desk & fake$_{plus}$ & 82.07 & 71.82 \\ \cline{2-5}
			& tote & fake$_{plus}$ & 82.69 & 76.84 \\ \cline{2-5}
			\multirow{3}{*}{Annotator 2} & shelf & fake$_{plus}$ & 65.62 & 41.48 \\ \cline{2-5}
			& desk & fake$_{plus}$ & 81.52 & 72.06 \\ \cline{2-5}
			& tote & fake$_{plus}$ & 82.14 & 75.94 \\ \cline{2-5}
            \multirow{3}{*}{Annotator 3} & shelf & fake$_{plus}$ & 62.18 & 51.61 \\ \cline{2-5}
			& desk & fake$_{plus}$ & 81.91 & 72.39 \\ \cline{2-5}
			& tote & fake$_{plus}$ & 79.02 & 64.49 \\ \cline{2-5}
            \multirow{3}{*}{Annotator 4} & shelf & fake$_{plus}$ & 66.22 & 57.03 \\ \cline{2-5}
			& desk & fake$_{plus}$ & 81.08 & 74.09 \\ \cline{2-5}
			& tote & fake$_{plus}$ & 81.78 & 70.37 \\ \cline{2-5}
            \multirow{3}{*}{Annotator 5} & shelf & fake$_{plus}$ & 63.27 & 52.53 \\ \cline{2-5}
			& desk & fake$_{plus}$ & 81.54 & 72.45 \\ \cline{2-5}
			& tote & fake$_{plus}$ & 82.89 & 73.27 \\ \cline{2-5}
            \multirow{3}{*}{Annotator 6} & shelf & fake$_{plus}$ & 66.05 & 46.90 \\ \cline{2-5}
			& desk & fake$_{plus}$ & 80.17 & 73.30 \\ \cline{2-5}
			& tote & fake$_{plus}$ & 78.12 & 63.76 \\ \cline{2-5}
            \multirow{3}{*}{Annotator 7} & shelf & fake$_{plus}$ & 65.33 & 50.15 \\ \cline{2-5}
			& desk & fake$_{plus}$ & 80.08 & 68.18 \\ \cline{2-5}
			& tote & fake$_{plus}$ & 81.94 & 70.27 \\ \cline{2-5}
            \multirow{3}{*}{Annotator 8} & shelf & fake$_{plus}$ & 66.37 & 53.67 \\ \cline{2-5}
			& desk & fake$_{plus}$ & 79.69 & 66.24 \\ \cline{2-5}
			& tote & fake$_{plus}$ & 82.74 & 64.20 \\ \cline{2-5}
            \multirow{3}{*}{Annotator 9} & shelf & fake$_{plus}$ & 62.51 & 52.48 \\ \cline{2-5}
			& desk & fake$_{plus}$ & 77.32 & 68.47 \\ \cline{2-5}
			& tote & fake$_{plus}$ & 82.35 & 70.71 \\ \cline{2-5}
            \multirow{3}{*}{Annotator 10} & shelf & fake$_{plus}$ & 64.03 & 57.93 \\ \cline{2-5}
			& desk & fake$_{plus}$ & 78.77 & 70.42 \\ \cline{2-5}
			& tote & fake$_{plus}$ & 77.89 & 69.11 \\ \cline{2-5}
            \multirow{3}{*}{Annotator 11} & shelf & fake$_{plus}$ & 66.14 & 52.13 \\ \cline{2-5}
			& desk & fake$_{plus}$ & 81.02 & 74.66 \\ \cline{2-5}
			& tote & fake$_{plus}$ & 80.36 & 68.57 \\ \cline{2-5}
            \multirow{3}{*}{Annotator 12} & shelf & fake$_{plus}$ & 68.45 & 49.90 \\ \cline{2-5}
			& desk & fake$_{plus}$ & 82.62 & 73.46 \\ \cline{2-5}
			& tote & fake$_{plus}$ & 81.80 & 68.15 \\ \cline{2-5}
            \multirow{3}{*}{Annotator 13} & shelf & fake$_{plus}$ & 66.23 & 56.17 \\ \cline{2-5}
			& desk & fake$_{plus}$ & 78.42 & 65.43 \\ \cline{2-5}
			& tote & fake$_{plus}$ & 79.42 & 68.80 \\ \cline{2-5}
            \multirow{3}{*}{Annotator 14} & shelf & fake$_{plus}$ & 64.16 & 51.81 \\ \cline{2-5}
			& desk & fake$_{plus}$ & 80.10 & 66.98 \\ \cline{2-5}
			& tote & fake$_{plus}$ & 79.65 & 65.01 \\ \cline{2-5}
            \multirow{3}{*}{Annotator 15} & shelf & fake$_{plus}$ & 66.39 & 49.60 \\ \cline{2-5}
			& desk & fake$_{plus}$ & 81.40 & 67.20 \\ \cline{2-5}
			& tote & fake$_{plus}$ & 82.51 & 71.63 \\ \cline{2-5}
            \multirow{3}{*}{Annotator 16} & shelf & fake$_{plus}$ & 66.21 & 54.94 \\ \cline{2-5}
			& desk & fake$_{plus}$ & 79.91 & 71.88 \\ \cline{2-5}
			& tote & fake$_{plus}$ & 81.83 & 72.17 \\ \cline{2-5}
            \multirow{3}{*}{Annotator 17} & shelf & fake$_{plus}$ & 65.46 & 51.92 \\ \cline{2-5}
			& desk & fake$_{plus}$ & 81.45 & 71.64 \\ \cline{2-5}
			& tote & fake$_{plus}$ & 82.19 & 69.23 \\ \cline{2-5}
            \multirow{3}{*}{Annotator 18} & shelf & fake$_{plus}$ & 61.83 & 52.75 \\ \cline{2-5}
			& desk & fake$_{plus}$ & 80.59 & 68.95 \\ \cline{2-5}
			& tote & fake$_{plus}$ & 77.71 & 63.43 \\ \cline{2-5}
            \multirow{3}{*}{Annotator 19} & shelf & fake$_{plus}$ & 69.15 & 51.08 \\ \cline{2-5}
			& desk & fake$_{plus}$ & 79.25 & 65.11 \\ \cline{2-5}
			& tote & fake$_{plus}$ & 81.74 & 67.75 \\ \cline{2-5}
            \multirow{3}{*}{Annotator 20} & shelf & fake$_{plus}$ & 66.78 & 57.10 \\ \cline{2-5}
			& desk & fake$_{plus}$ & 76.81 & 68.52 \\ \cline{2-5}
			& tote & fake$_{plus}$ & 82.27 & 69.93 \\ \bottomrule
		\caption{Results of instance segmentation tasks where training samples generated by priors of 20 annotators. Segmentation tasks are conducted by Mask R-CNN.}
		\protect\label{table:know}
        \end{longtable}

To make sure the diversity, we summoned 20 people with different ages, genders, nationalities to annotate the objects parallel, and then generated the scene layout accordingly.

The reason why different priors lead to insignificant difference, as we assume, is that despite that different people have different preferences on how to place an object in a given scene, they agree on what pose, location or relationship is possible. There are also extreme cases when one person thinks one form of placement never happens, and others think not (i.e. whether a drink bottle can be placed upside down). But in general, they achieved an agreement unconsciously. Thus, the distribution of layouts generated under diverse preferences are not very different. Another reason might be that current CNN networks are proved to have sufficient capabilities to cope with these minor differences of pose, location and relationship. And at the same time, some data augmentation techniques incorporated default can also facilitate to further reduce the impacts.

\subsection{GAN Refined Results}
\begin{figure}[!ht]
	\begin{center}
		\includegraphics[width=9cm,height=12cm]{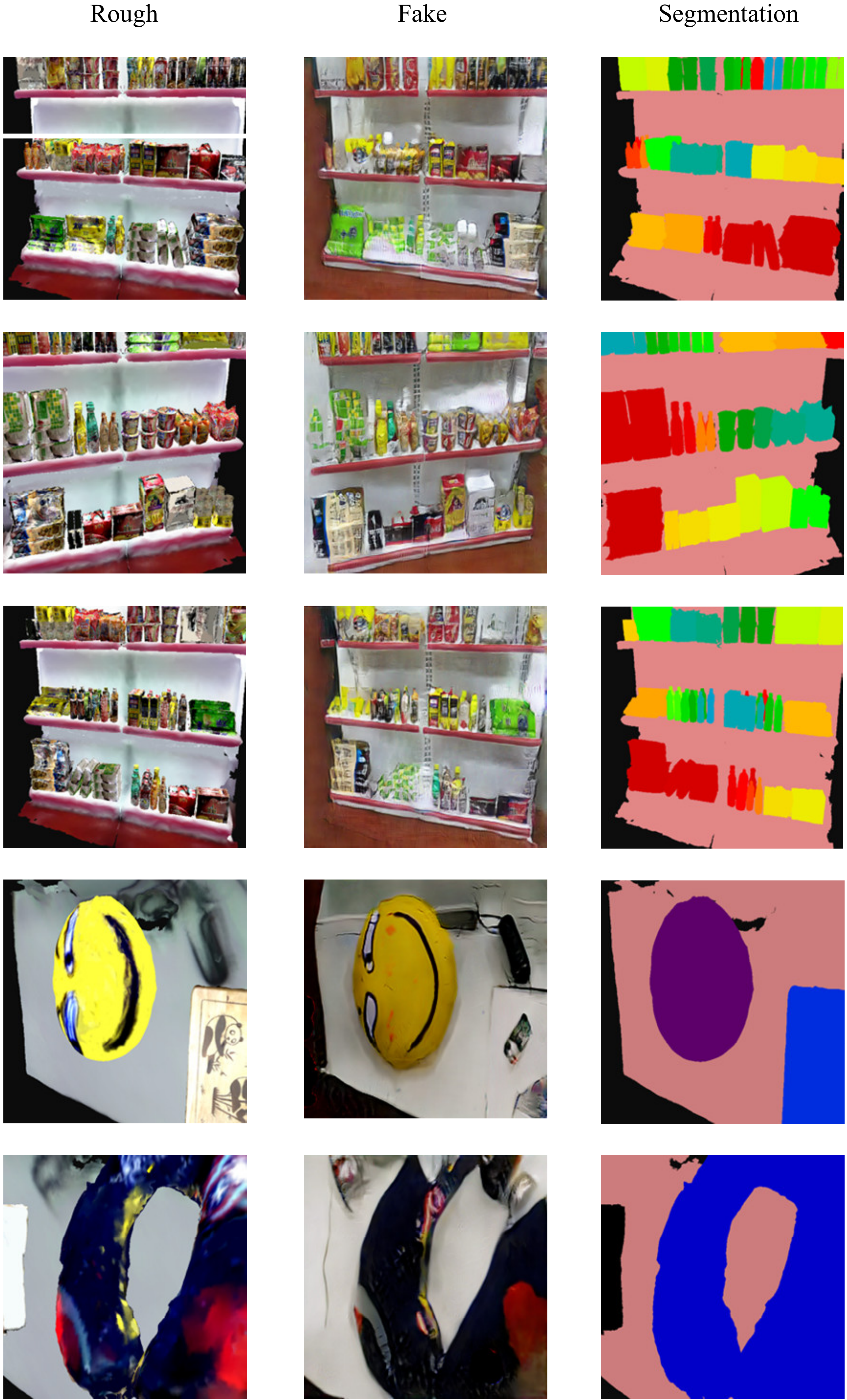}
	\end{center}
	\caption{More GAN refined results. They are naturally paired with pixel-wise accuracy segmentations.}\protect\label{fig:refineShelf}
\end{figure}

\begin{figure}[!ht]
	\begin{center}
		\includegraphics[width=9cm,height=8cm]{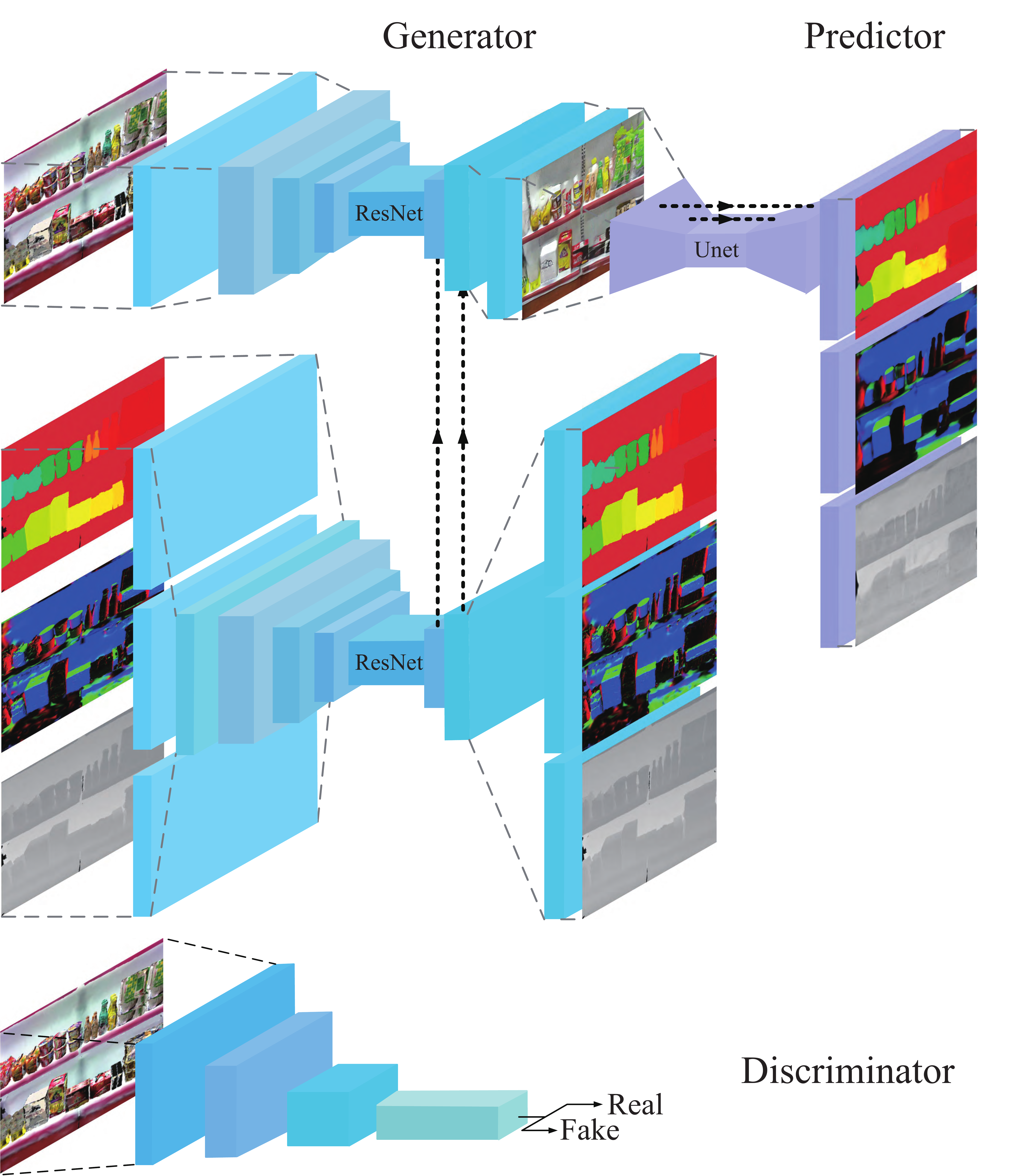}
	\end{center}
	\caption{Visualization of GeoGAN architecture.}\protect\label{fig:GAN}
\end{figure}

More refined results are listed in Figure \ref{fig:refineShelf}. And Figure \ref{fig:GAN} shows more visualized details of the GeoGAN structure. 

\end{document}